\RequirePackage[2018-12-01]{latexrelease}

\documentclass{tADR2e}

\begin{document}

\jvol{33} \jnum{12} \jyear{2019} \jmonth{May}

\title{Mobile Robot Localization with GNSS Multipath Detection using Pseudorange Residuals}

\author{Taro Suzuki$^{a}$$^{\ast}$\thanks{$^\ast$Corresponding author. Email: taro@furo.org
\vspace{6pt}}
\\\vspace{6pt} 
$^{a}${\em{Future Robotics Technology Center, Chiba Institute of Technology, Japan}};
\\\vspace{6pt}
\received{v1.0 released May 2019} }

\maketitle

\begin{abstract}
  This paper proposes a novel positioning technique suitable for use in mobile robots and vehicles in urban environments in which large global navigation satellite system (GNSS) positioning errors occur because of multipath signals. During GNSS positioning, the GNSS satellites that are obstructed by buildings emit reflection and diffraction signals, which are called non-line-of-sight (NLOS) multipath signals. These multipath signals cause major positioning errors. To mitigate the NLOS multipath errors, the NLOS signals must be identified from all of the received GNSS signals. The key concept considered in this paper is the estimation of a user’s position using the likelihood of the position hypotheses computed from the GNSS pseudoranges, consisting only of LOS signals based on the analysis of the pseudorange residuals. To determine the NLOS GNSS signals from the pseudorange residuals at the user’s position, it is necessary to accurately determine the position before the computation of the pseudorange residuals. This problem is solved using a particle filter. We propose a likelihood estimation method using the Mahalanobis distance between the hypotheses of the user’s position computed from only the LOS pseudoranges and the particles. To confirm the effectiveness of the proposed technique, a positioning test was performed in a real-world urban environment. The results demonstrated that the proposed method is effective for accurately estimating the user’s position in urban canyons, where conventional GNSS positioning can lead to large positioning errors.
    
\medskip

\end{abstract}

\section{Introduction}
Global navigation satellite systems (GNSSs) are gaining increasing popularity owing to their wide range of current and potential applications. Significant improvements in the availability of satellite-based positioning in urban areas are expected because of an increase in the number of positioning satellites launched by various countries. Navigation signals from these satellites are expected to be used in urban contexts for applications, such as intelligent transportation systems, automatic vehicle navigation, and mobile robot navigation. However, because of the serious impacts on the positioning accuracy of multipath signals from a positioning satellite caused by urban canyon environments, an increased availability of satellite positioning techniques does not necessarily equate to more precise positioning. As multipath effects are highly dependent on the surrounding geographic features, such as tall buildings near GNSS receivers, errors cannot be eliminated by the use of differential GNSSs. Figure 1 shows the multipath signals in an urban environment. A GNSS multipath signal can be divided into two types: line-of-sight (LOS) and non-line-of-sight (NLOS) multipath signals. For the LOS multipath signal case, both direct and reflected/diffracted signals are simultaneously received. This affects the GNSS positioning accuracy \cite{1}. However, various signal correlation techniques, such as narrow \cite{2} and strobe \cite{3} correlators, can be used to mitigate the LOS multipath errors \cite{2,3,4,5}. At most, the LOS multipath error is several meters. For the NLOS multipath signal case, the GNSS receiver can only receive multipath signals that are reflected off the surfaces of buildings or diffracted by their edges from a satellite behind the buildings, and the multipath signal will have a large error range of several tens of meters or more in the GNSS pseudorange measurements. Hence, practical techniques for NLOS signal detection are required to improve positioning accuracies in urban environments.

\begin{figure}[t]
  \centering
  \includegraphics[width=125mm]{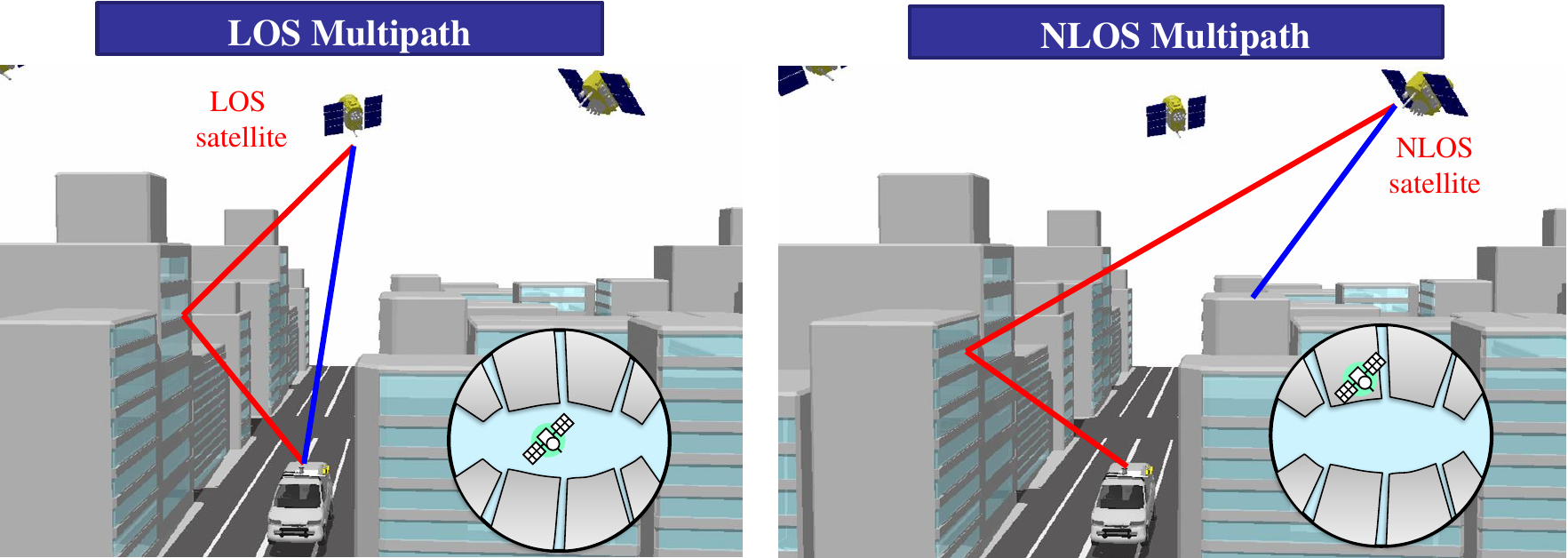} 
  \caption{LOS and NLOS multipath signals in urban environments. NLOS multipath significantly affects the positioning accuracy.}
  \label{fig:1}
\end{figure}

To mitigate the NLOS multipath error, NLOS signals (or invisible satellites) must be identified from all received GNSS signals. In general, it is difficult to identify NLOS signals from the received signals alone. Additional sensors are required to detect NLOS signals with GNSS receivers. Marais et al. proposed an NLOS detection tool using an image sensor \cite{6}. However, NLOS detection is affected by the weather and illumination conditions. We previously proposed the use of a fish-eye camera and an image processing technique to detect NLOS signals \cite{7}. This technique is more robust than the previous method, however techniques that use visible-light cameras cannot be used during the nighttime. In some of our other previous works, we proposed a technique to realize NLOS signal detection using an omnidirectional far-infrared (FIR) camera to exclude invisible satellites \cite{8, 9}. Using an FIR camera, it is easy to distinguish the sky from obstacles such as buildings. It can also be used during the nighttime. However, omnidirectional FIR cameras are not readily available on the market and are only produced upon special request. Maier et al. proposed the use of real-time range information from obstacles using a laser scanner to identify invisible satellites with associated NLOS multipath errors \cite{10}. However, it is difficult to apply this technique to moving vehicles to obtain their real-time position.

In previous studies, 3D environmental models were used to predict LOS and NLOS signals for the purpose of improving the GNSS positioning accuracy \cite{12,22}. Several 3D mapping-aided GNSS positioning algorithms were developed by several research groups. A technique called GNSS shadow matching \cite{15,16,17}, which uses 3D city models to estimate the user position based on the degree of correlation between the predicted and measured GNSS satellite visibility, has been proposed. Some researchers have proposed other techniques that are derived from GNSS shadow matching \cite{11, 14, 18, 19}. We also proposed a method that uses signal propagation simulations in 3D maps for actual positioning by comparing the simulated signal strength with that of the actual signal received \cite{20}. We proposed a positioning method that can directly combine GNSS pseudorange-based positioning and 3D environmental maps to improve the positioning accuracy in urban environments \cite{21}. This method can estimate the user’s position using the likelihood of position hypotheses computed from GNSS pseudoranges, which consist of only LOS signals determined from the 3D map. These 3D mapping-aided GNSS positioning algorithms can improve the positioning accuracy in urban environments, however, problems relating to the system complexity and costs of maintaining a 3D map negatively impact their use. 

In this study, we propose a novel GNSS positioning method that uses a particle filter (PF) with NLOS GNSS signal detection, without requiring 3D maps. This study was inspired by the work and ideas presented in Refs. \cite{23, 24}. In these studies, a 3D map was used to determine the LOS and NLOS signals, and then the likelihood of the position candidate was computed from the measurement residuals. Our method does not require a 3D map to detect the NLOS signals. Instead, the measurement residuals are directly used to generate the subset of the LOS measurement hypothesis. In addition, we use a particle filter to estimate the position of moving objects, such as vehicles or mobile robots, based on the likelihood of particles, which is computed from the Mahalanobis distance between the particle location and GNSS position solution using the subset of the LOS measurement hypothesis.

\begin{figure}[!t]
  \centering
  \includegraphics[width=155mm]{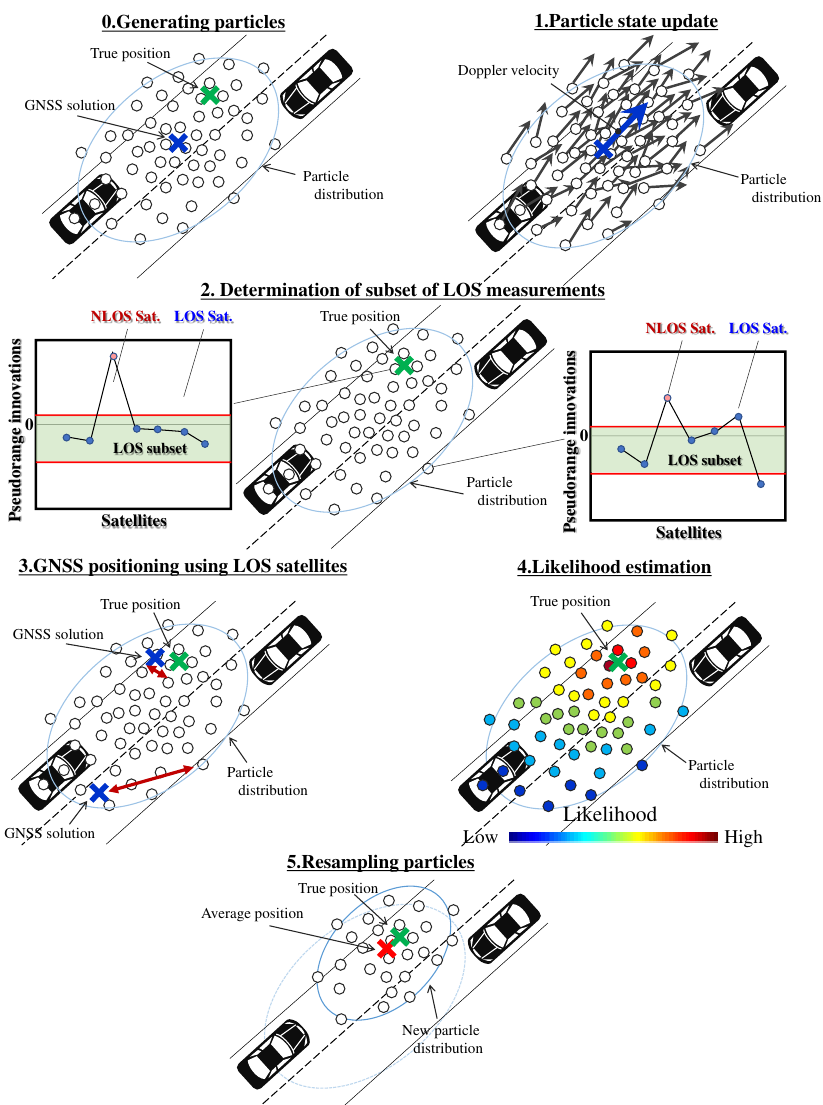} 
  \caption{Overview of the proposed positioning method using a particle filter. NLOS GNSS signals are eliminated based on pseudorange residuals.}
  \label{fig:2}
\end{figure}

\section{Proposed method overview}
The key idea behind this work is to estimate the user position by employing the likelihood-of-position hypotheses computed from the GNSS pseudoranges, consisting only of the LOS signals estimations based on the pseudorange residuals. The NLOS multipath error is always a positive value because the NLOS signals are delayed, compared with the direct signals. We exploit this phenomenon to exclude the NLOS signals and select LOS signals based on the pseudorange residuals. 

To compute the pseudorange residuals to determine the LOS GNSS signals at the user’s position, it is in turn necessary to accurately determine the position before the residual computation; we solve this problem using a particle filter, which provides a numerical approximation of the estimated states of a set of weighted random samples, called particles. Differing hypotheses of user position are represented by these particles and, for each of these, their likelihood can be evaluated using only the LOS pseudoranges determined from the pseudorange residuals.

Figure 2 illustrates the proposed positioning algorithm that employs the particle filter. The proposed algorithm works as follows:

\begin{enumerate}
  \item The hypothetical user positions are created as particles of which the state vectors have 3D positions; x, y, and z in Cartesian earth-centered earth-fixed (ECEF) coordinates. The initial particles are distributed based on the solution of the normal GNSS point positioning and its variance. 
  \item In the prediction step of the particle filter, the particle state in the next time step is estimated using a motion model. We consider a motion model based on the velocity of each particle computed from GNSS Doppler measurements. Here, the initial velocity is also computed via normal GNSS point positioning using all of the GNSS measurements.
  \item Next, we compute the measurement residuals at each particle location. First, the geometric distance between each satellite and particle is computed and compensated for. To eliminate the receiver clock error, the difference in the pseudoranges across satellites is calculated. As a result, the multipath and unmodeled errors remain if the particle is close to the true position. The magnitude of the NLOS multipath error is generally greater than that of other errors, and correct NLOS measurements can be determined if the particle is located near the true position. We generate the subset of the LOS measurement hypothesis at each particle using a simple threshold test. 
  \item Then, we compute the GNSS position solution using the generated subset of the LOS measurement hypothesis at each particle location. The position solution can be expected to move toward the true position if the particle location is close to the true position because the NLOS signals are eliminated. In contrast, the new position solution for a particle that is far from the true position does not move toward the true position because an incorrect subset of the LOS measurements is used. In other words, the distance between the particle location and new position solution indicates the likelihood that the particle represents the true location.
  \item The likelihood of each particle is computed from the Mahalanobis distance between the particle location and new GNSS position solution. The particle that is closest to the true position has a higher likelihood compared with those of other particles. The probabilistic weight of each particle is then updated based on its likelihood. 
  \item Finally, all of the particles are resampled based on their new weights. Particles that are sufficiently close to the true position survive this resampling step, and the average position of the remaining particles is employed as the final estimated user position.
\end{enumerate}

The remainder of this paper describes the methods for the computation of the pseudorange residuals and detection of the NLOS signals from the pseudorange residuals. Next, we demonstrate how the particle likelihoods are obtained from the Mahalanobis distance. The performance of our proposed method is then evaluated by estimating the positions in actual urban canyon environments.

\section{NLOS detection from pseudorange residuals}
\subsection{Pseudorange residuals computation}
To exclude the NLOS signals, we compute the pseudorange residuals at each particle position. Hypothetical user positions are created as particles of which the state vectors have 3D positions; $x$, $y$, and $z$ in Cartesian ECEF coordinates. The initial particles are distributed based on the solution of the normal GNSS point positioning and its variance. Here, the state of the $i$-th particle is denoted as follows.

\begin{equation}
  \mathbf{x}_i=\left[\begin{array}{lll}
  x & y & z
  \end{array}\right]^T
\end{equation}

\noindent The pseudorange of satellite $j$ is denoted as follows.

\begin{equation}
  \rho^j=r^j+c\left(d t-d T^j\right)+I^j+T^j+\varepsilon^j
\end{equation}

In the above equation, $r$ is the geometric satellite-to-receiver distance, $I$ is the ionospheric delay, $T$ is the tropospheric delay, $c$ is the velocity of light, $dt$ is the clock bias of the receiver, $dT$ is the clock bias of the satellite, and $\varepsilon$ is the measurement error, including the NLOS multipath error. The clock bias $dT$ of the satellite can be calculated from the broadcasted clock error information. The tropospheric delay $T$ is compensated using the Saastamoinen and Hopfield models. Here, the ionospheric delay $I$ is compensated using the Klobuchar model. The compensated pseudorange $\tilde{\rho}$ is denoted as follows

\begin{equation}
  \tilde{\rho}^j=r^j+c d t+\varepsilon^j
\end{equation}

\noindent where the geometric distance between each satellite $j$ and each particle $i$ is computed by

\begin{equation}
  r_i^j \approx\left\|\mathbf{C} \mathbf{x}^j-\mathbf{x}_i\right\|
\end{equation}

\noindent where $\mathbf{x}^j$ is the satellite position in ECEF coordinates, and $\mathbf{C}$ is the rotation matrix used to compensate for the Sagnac effect. To eliminate the receiver clock error, the difference in the pseudoranges across satellites is calculated. A reference satellite is required to compute the difference in the pseudoranges across satellites. Here, a high-elevation satellite with a strong signal-to-noise ratio (SNR) is used as the reference satellite $k$. We assume that this reference satellite is not an NLOS satellite and that the pseudorange does not contain a large multipath error. Therefore, the pseudorange residuals at the $i$-th particle, $\delta z_i^j$, are denoted by

\begin{equation}
  \begin{gathered}
  \delta z_i^j=\left(\tilde{\rho}^j-r_i^j\right)-\left(\tilde{\rho}^k-r_i^k\right) \\
  =\varepsilon^j-\varepsilon^k \\
  \approx \varepsilon^j
  \end{gathered}
\end{equation}

\noindent where, $j \neq k$. In Equation (5), the receiver clock error can be eliminated by calculating the difference between the pseudoranges. The multipath error in the reference satellite $k$ can be assumed to be almost zero. As a result, the multipath and unmodeled errors in the pseudorange $j$ remain in Equation (5) if the particle is close to the true position. In contrast, an incorrect geometric distance $r_i^j$ is computed for particles that are far from the true position. Thus, incorrect pseudorange residuals are calculated for “incorrect” particles. It is assumed that the NLOS signals can be detected at the true position using the correct pseudorange residuals. We detect NLOS signals and generate the hypothesis for the LOS signal subsets at each particle location.

\subsection{NLOS detection}
We propose a method to exclude the NLOS signals from the pseudorange residuals. The magnitude of NLOS multipath error is generally greater than that of other errors, and correct NLOS measurements can be easily determined from their residuals if the associated particle is located near the true position. We generate the subset of the LOS measurement hypothesis at each particle using a simple threshold test. The subset of the LOS measurements at the $i$-th particle $L_i$ is denoted by

\begin{equation}
  L_i=\left\{\tilde{\rho}^j \mid T_L \leq \delta z_i^j \leq T_U\right\}
\end{equation}

\noindent where, $T_L$ and $T_U$ are the upper and lower thresholds to select LOS measurements from the measurement residuals, respectively. As previously mentioned, the NLOS multipath error is always a positive value, and thus we chose thresholds to exclude the positive residuals as follows.

\begin{equation}
  \left|T_L\right|>\left|T_U\right|, T_L<0, T_U>0
\end{equation}

Using this threshold test, it is assumed that the correct NLOS signals are excluded and that the correct subsets of the LOS measurements are generated at particles near the true position. In contrast, for particles far from the true position, an incorrect subset of the LOS measurements hypothesis will be generated. The likelihood of each particle is computed from the positioning solution obtained using this subset of the LOS measurement hypothesis.

\section{Particle filter with NLOS detection}
To estimate the user’s position, we employ probabilistic localization, which is commonly known as a particle filter, following a recursive Bayesian filtering scheme. Particle filters provide a numerical approximation of estimated states based on a set of weighted random samples. The user’s position can be hypothesized as a particle in this particle filter method, and the computation of the pseudorange residuals can be employed to estimate the user’s position. 

The particle filter has two steps: prediction and observation. During the observation step, the set of particles is sampled with probabilities proportional to the weights assigned by the observation model. A weight is assigned to each particle based on the likelihood model of each particle, and the particle set is then resampled according to these assigned weights to obtain a better position distribution approximation. The framework and implementation of particle filters have been reported previously \cite{25}, and therefore only descriptions of the designs of the motion and observation models for the particle filter are provided in this section.

\subsection{Prediction step}
In the particle filter prediction step, the particle state in the next time step is estimated using a motion model. We consider a motion model based on the velocity of each particle computed from GNSS Doppler measurements. The particle state of the next time step is updated using the particle’s velocity calculated as

\begin{equation}
  \mathbf{x}_{i, t}=\mathbf{x}_{i, t-1}+\mathbf{v}_{i, t-1} \Delta t+\mathbf{w}
\end{equation}

\noindent Here, $\Delta t$ is the time step and $\mathbf{v}$ is the 3D velocity computed only from the LOS Doppler measurements determined in the previous subset of LOS measurements at each particle location. $\mathbf{w}$ is the added noise vector with a Gaussian distribution and is given by

\begin{equation}
  p\left(w\right)=\frac{1}{\sqrt{2\pi}\sigma}exp\left(-\frac{1}{2\sigma}w\right)
\end{equation}

\noindent Here, the dispersion standard deviation, $\sigma$, is based on experimental values derived from positioning tests. In addition, the initial velocity is computed from the results of the normal GNSS point positioning using all of the GNSS measurements.

\subsection{Observation step}
The observation step samples the set of particles with probabilities proportional to the weights assigned by the observation model. An observation model can be designed that is suitable for a given test environment. In this study, we use the distance between the particle location and new GNSS position solution computed from the subset of the LOS measurement hypothesis. The position solution can be expected to move toward the true position if the particle location is close to the true position because the NLOS signals are eliminated. In contrast, the new position solution for the particles that are far from the true position does not move toward the true position because incorrect subsets of the LOS measurements are used. 

The likelihood of each particle is computed from the Mahalanobis distance between the particle location and new GNSS position solution, which is greater if the particle is further from the true position. Here, the Euclid distance between the particle and new GNSS position solution will also be greater, similar to the Mahalanobis distance. However, in urban environments, the dilution of precision (DOP) is greater because of the limited number of available GNSS satellites. The DOP represents the contribution of the satellite geometry to the positioning accuracy. If the position is computed with a large DOP value because of poor satellite geometry, the positioning accuracy is low, even when using only LOS signals. In such cases, the GNSS positioning error distribution is biased toward a fixed direction. For this reason, the Euclidian distance does not correctly represent the likelihood of the particles, and the Mahalanobis distance should be used.

To compute the Mahalanobis distance, the new GNSS position solution is calculated from the results of single point positioning (SPP) using the subset of the LOS measurement hypothesis at the ith particle position. The new GNSS position solution is denoted by

\begin{equation}
  \mathbf{\mu}_i=\left[\begin{array}{lll}
    x_{\mathrm{gnss},i} & y_\mathrm{{gnss},i} & z_{\mathrm{gnss},i}
  \end{array}\right]^T
\end{equation}

\noindent The new GNSS position solution will be computed repeatedly for the number of particles. The Mahalanobis distance $D_i$ between the ith particle position and GNSS position solution is calculated as follows: 

\begin{equation}
  D_i=\sqrt{\left(\mathbf{x}_i-\mathbf{\mu}_i\right)^T\Sigma_i^{-1}\left(\mathbf{x}_i-\mathbf{\mu}_i\right)}
\end{equation}

\noindent where $\Sigma_i$ is the GNSS position error covariance calculated from the results of single-point positioning using the subset of the LOS measurement hypothesis at the $i-th$ particle position. The likelihood of particle $p_i$ can be calculated as follows: 

\begin{equation}
  p_i\approx\frac{1}{\sqrt{2\pi}\sigma}exp\left(-\frac{D_i^2}{2\sigma^2}\right)
\end{equation}

\noindent Here, the standard deviation, $\sigma$, is based on experimental values derived from the positioning tests. Note that the likelihood of the particle in Equation (12) is maximized when the distance in Equation (11) is minimized, i.e., when the particle position completely matches the new GNSS position solutions using only LOS signals. 

A feature of the proposed likelihood estimation method is that it can also be used in open-sky environments. If there are no obstacles, the new GNSS position solution will be generated at a point close to the true position in each particle. Thus, a short Mahalanobis distance indicates particles that are close to the true position. In contrast, a long Mahalanobis distance indicates particles that are far from the true position.

\begin{figure}[t]
  \centering
  \includegraphics[width=100mm]{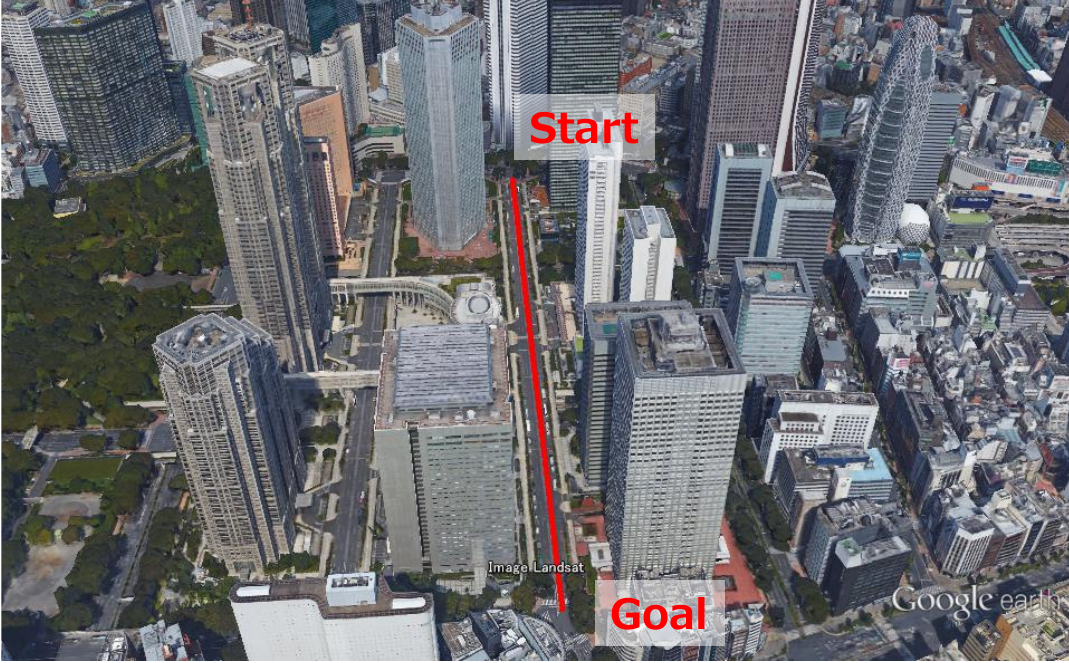} 
  \caption{Test course of the kinematic experiment in an urban canyon to evaluate the proposed positioning accuracy.}
  \label{fig:3}
\end{figure}

\begin{figure}[t]
  \centering
  \includegraphics[width=65mm]{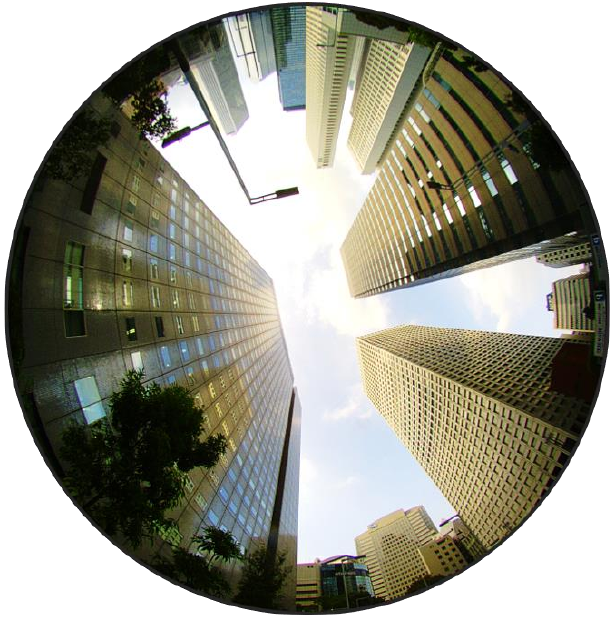} 
  \caption{Fish-eye image of the test environment. The environment is surrounded by tall buildings.}
  \label{fig:4}
\end{figure}

\section{Evaluation}
To confirm the effectiveness of the proposed technique, a kinematic positioning test was performed in Shinjuku, Tokyo, Japan. The environment and travel route are shown in Figure 3. The surroundings of the travel route consisted of scattered tall buildings (maximum height of 220 m), providing an environment susceptible to satellite masking. The fish-eye image captured at the GNSS antenna location during the test is shown in Figure 4. The test environment is surrounded by tall buildings. 

A GNSS antenna (Trimble Zephyr 2) and GNSS receiver (Trimble NetR9) were used to collect the GNSS data, which was obtained at a rate of 10 Hz. L1 pseudorange measurements of GPS, GLONASS, BeiDou, and QZSS were used in this experiment. The observation vehicle traveled approximately 500 m at speeds of less than 60 km/h. For our evaluations, a position and orientation estimation system for the land vehicles (POSLV, Applanix), with a multiple-frequency GNSS receiver and high-grade inertial navigation system, was used to estimate the reference positions used as the ground truth. 

First, we verified that many NLOS multipath signals were received from invisible satellites in this test. Using the ground truth, we computed the proposed pseudorange residuals and generated the subset of the LOS measurements at the true position. Figure 5 shows the pseudorange residuals of each satellite at the true position. The red solid lines in Figure 5 indicate the upper and lower thresholds, used to determine the LOS/NLOS measurements. From Figure 5, it can be seen that the pseudorange residuals sometimes increased in the positive direction because of the NLOS signal reception. We computed the proposed pseudorange residuals for an incorrect location. Figure 6 shows the pseudorange residuals at the incorrect position (the true position plus 10 m). Compared with Figure 5, some LOS measurements were excluded as NLOS signals and incorrect subsets of the LOS measurement hypothesis were generated.

The normal GNSS point positioning results are shown in Figure 7. The red points and gray ellipses in Figure 7 indicate the solutions of the normal GNSS point positioning and their covariance, respectively. The green line indicates the ground truth. These results demonstrate that the GNSS point positioning sometimes contained large errors due to the NLOS multipath signals. In addition, from the close-up of the plot in Figure 7, incorrect position error covariances were estimated due to NLOS multipath errors. The position solutions obtained using the NLOS exclusion method at the true position is shown in Figure 8. Large multipath errors were eliminated by using the proposed NLOS exclusion method. In addition, correct position error covariances were estimated. From Figure 8, it can be expected that the Mahalanobis distance will be short if the particle is close to the true position.

The Mahalanobis distance between the particle location and GNSS position solution using the subset of LOS measurement is shown in Figure 9. The blue and red lines in Figure 9 indicate the Mahalanobis distances at the true location and 10 m from the true location. At the true location, the Mahalanobis distance was less than that of the incorrect location. The position likelihood can be calculated from the Mahalanobis distance between the particle location and positioning solution using the subset of LOS measurements at the particle location.

\begin{figure}[!t]
  \centering
  \includegraphics[width=160mm]{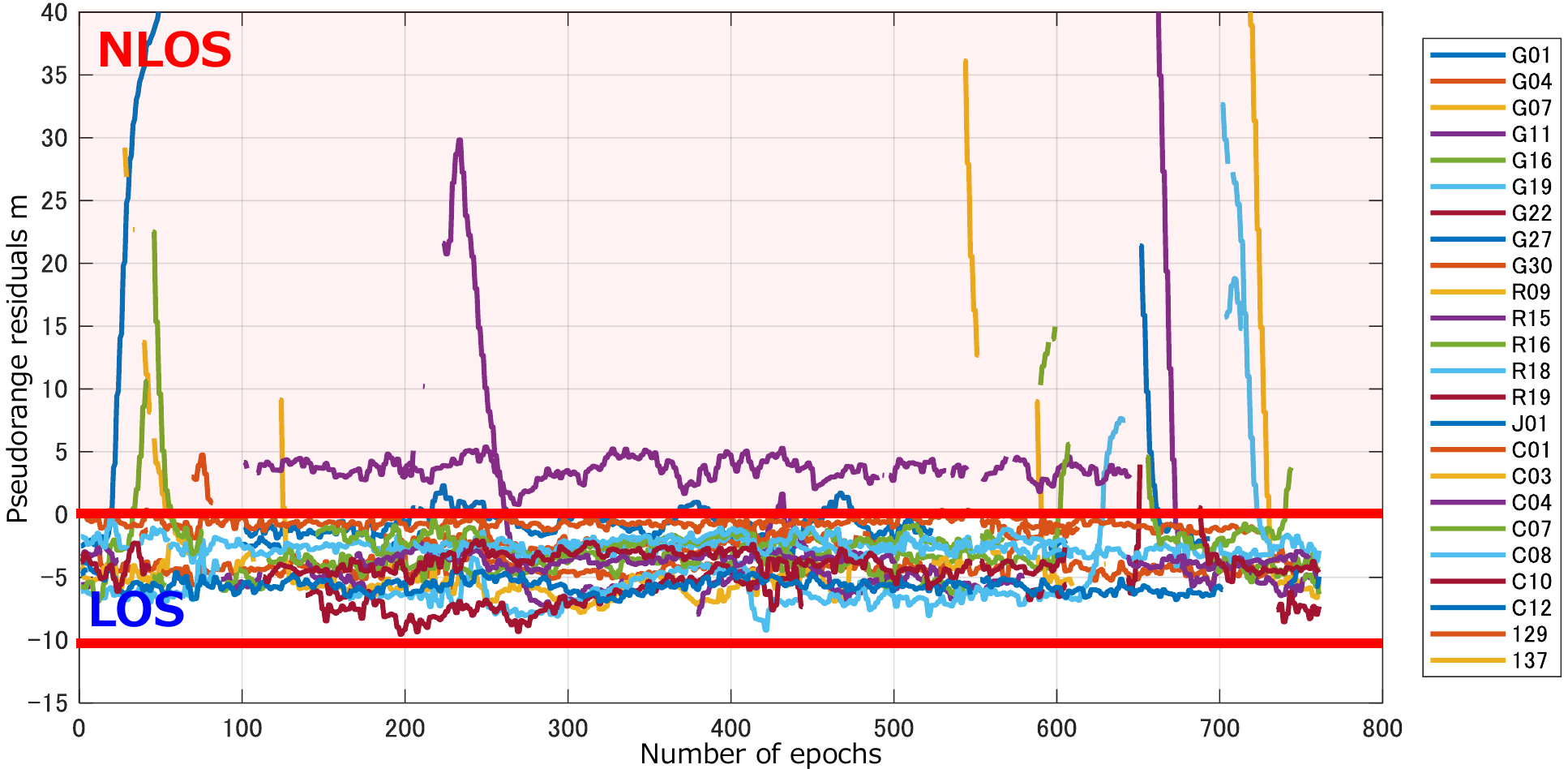} 
  \caption{Pseudorange residuals computed at the true position. Red lines indicate the thresholds to eliminate the large pseudorange residuals as NLOS signals.}
  \label{fig:5}
\end{figure}

\begin{figure}[!t]
  \centering
  \includegraphics[width=160mm]{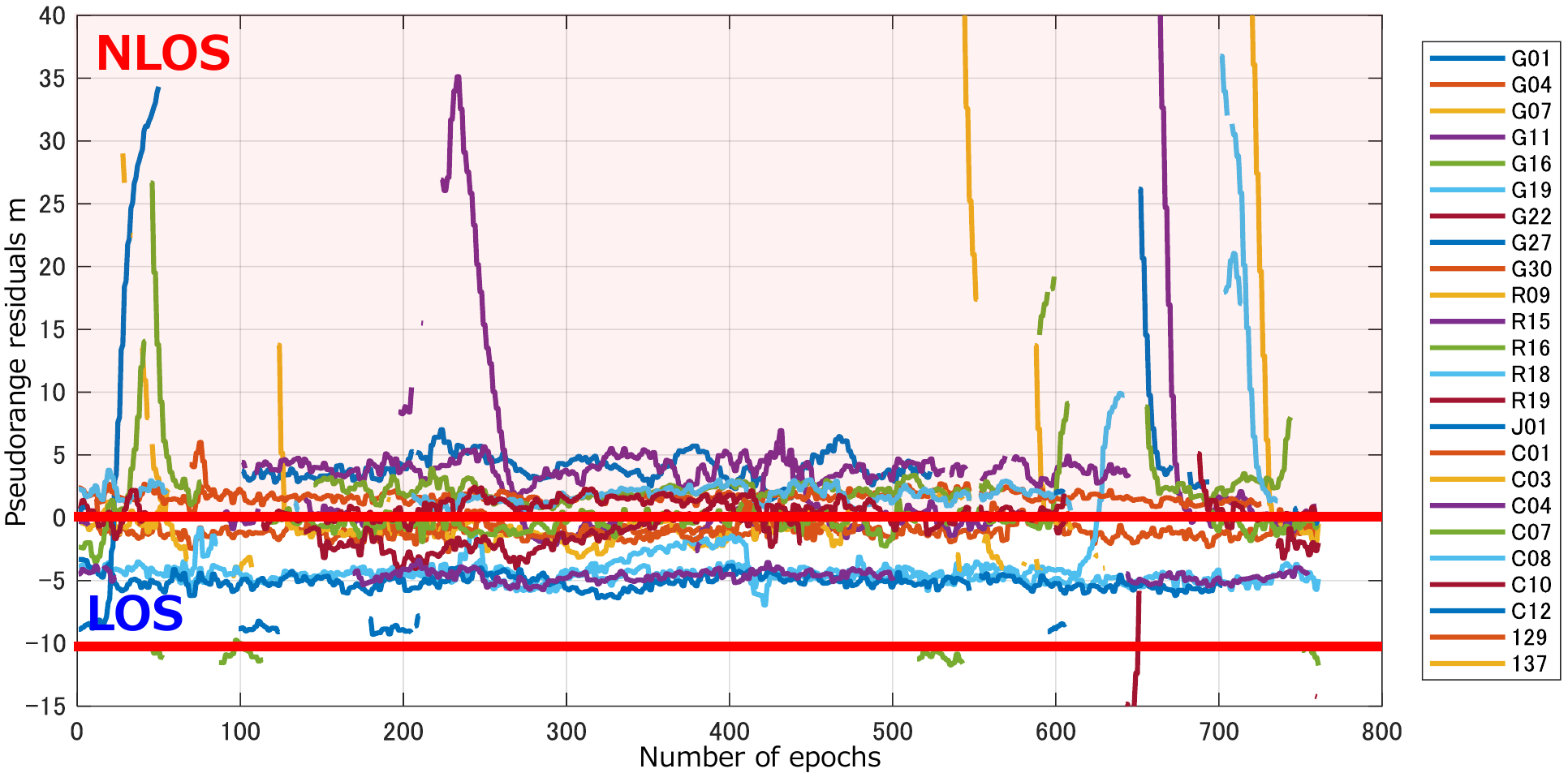} 
  \caption{Pseudorange residuals at an incorrect position (the true position plus 10 m). Compared with Figure 5, LOS signals is detected as NLOS signals and vice versa.}
  \label{fig:6}
\end{figure}

\clearpage

\begin{figure}[!t]
  \centering
  \includegraphics[width=82mm]{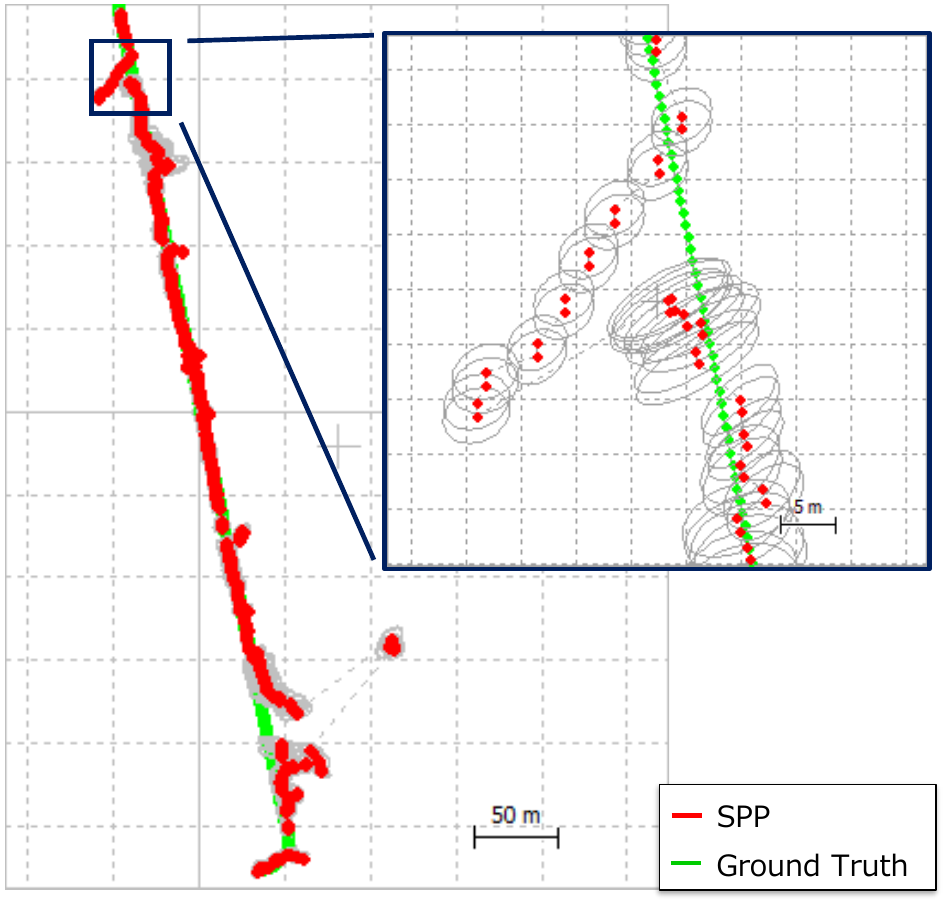} 
  \caption{Results of the GNSS single-point positioning without NLOS exclusion. GNSS position error covariance cannot be correctly estimated.}
  \label{fig:7}
\end{figure}

\begin{figure}[!t]
  \centering
  \includegraphics[width=82mm]{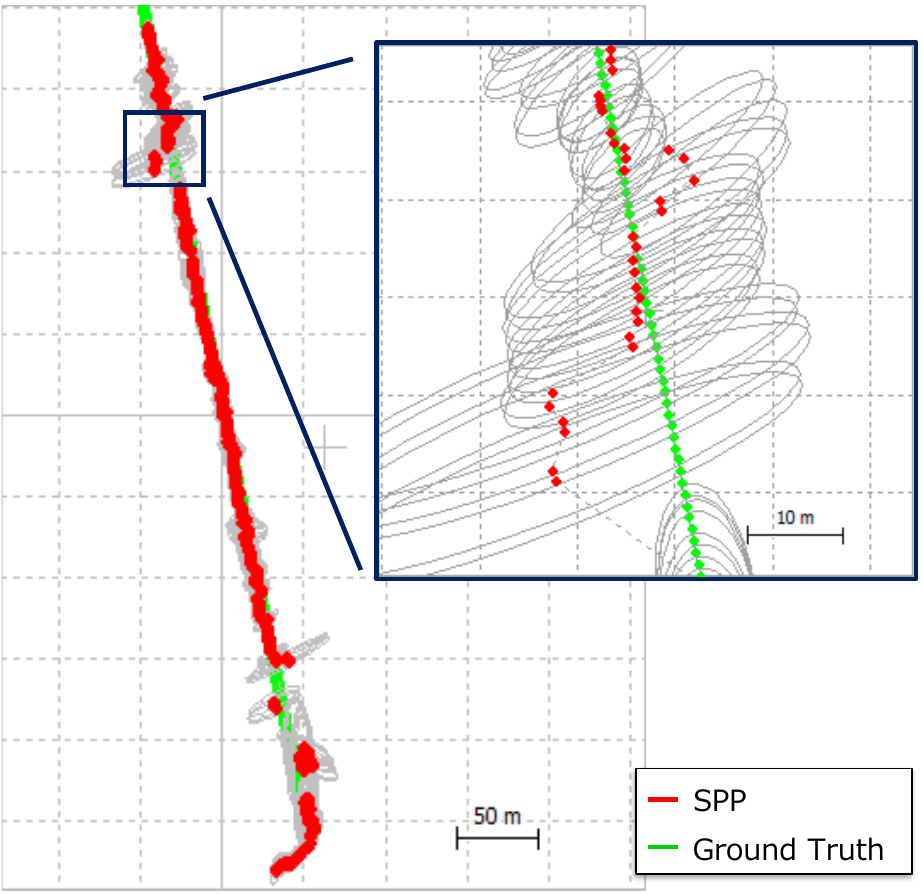} 
  \caption{Results of the GNSS single-point positioning with NLOS exclusion. Correct position error covariance is estimated because of the  NLOS exclusion.}
  \label{fig:8}
\end{figure}

\begin{figure}[!t]
  \centering
  \includegraphics[width=120mm]{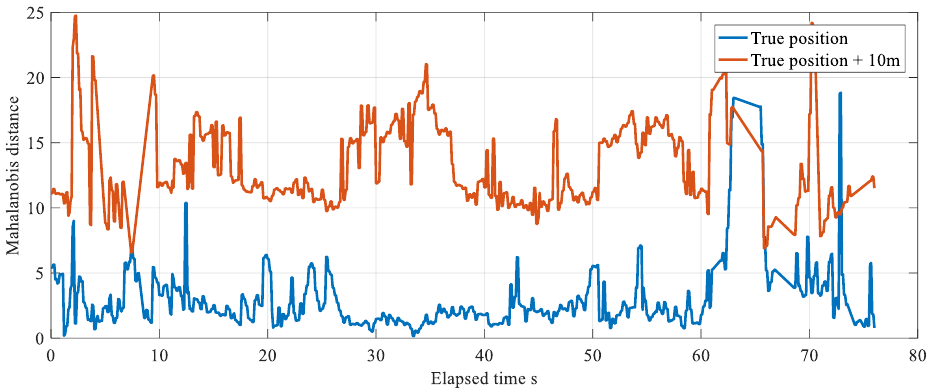} 
  \caption{Mahalanobis distance at the true and incorrect positions. At true position, Mahalanobis distance becomes short.}
  \label{fig:9}
\end{figure}

\clearpage

\begin{figure}[!t]
  \centering
  \includegraphics[width=150mm]{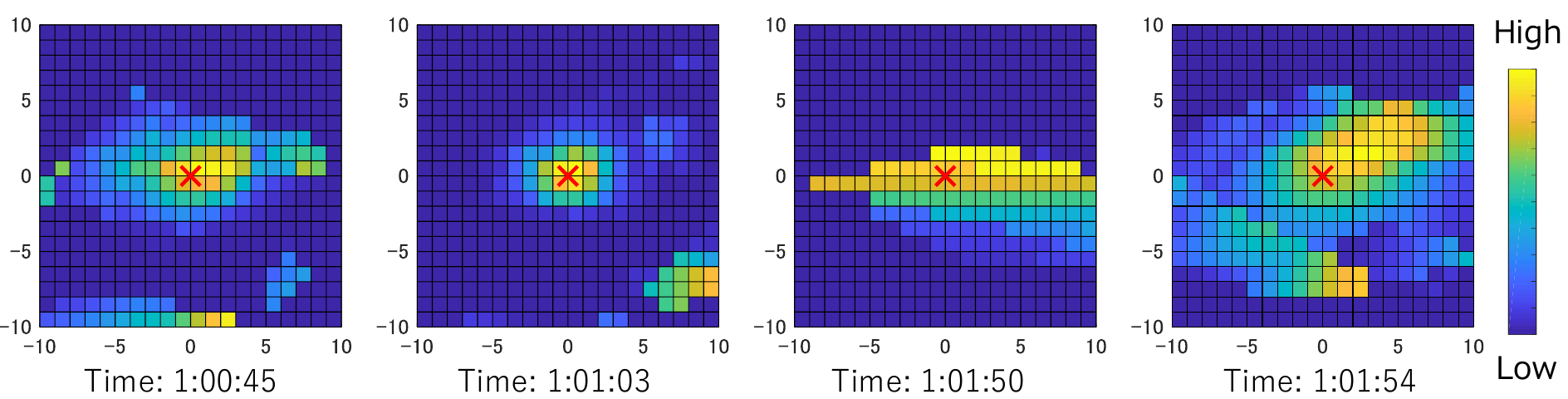} 
  \caption{Results of the likelihood estimation using the proposed method. High likelihoods are appeared at true position.}
  \label{fig:10}
\end{figure}

Figure 10 illustrates the likelihoods of the proposed method in a 1-m mesh grid point at different times. The red cross markers in Figure 10 indicate the true position. The yellow shading indicates a high likelihood, calculated using Equation (12). The likelihood was maximized at the true position using the proposed method. The distribution of high likelihoods is concentrated in the central area, and the likelihood distribution converged using only single-epoch observation data. These results indicate that the use of a particle filter is a robust method for estimating the true position.

\begin{figure}[!t]
  \centering
  \includegraphics[width=80mm]{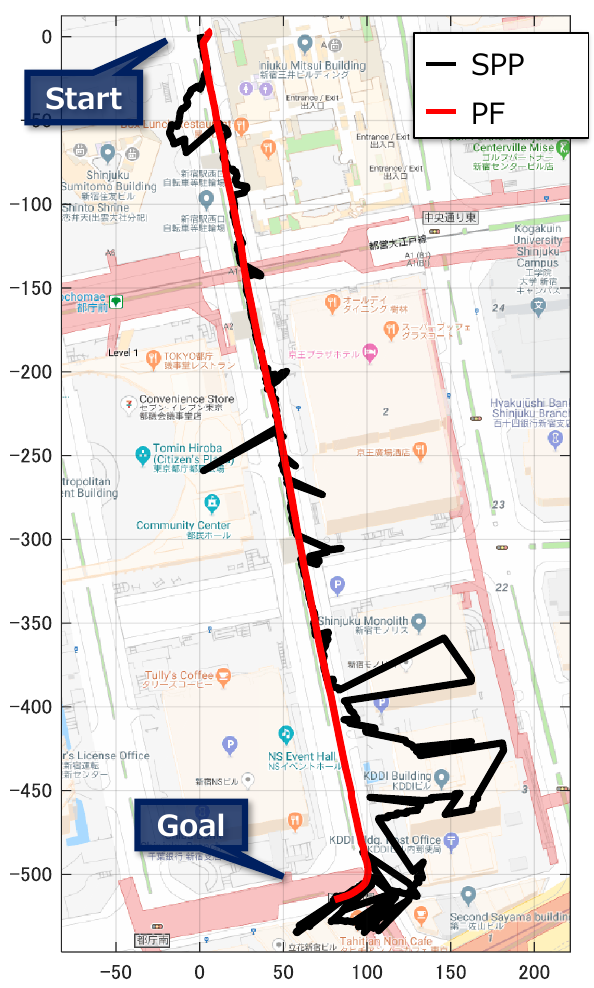} 
  \caption{Results of the position estimation using the particle filter. The proposed method (red line) can estimate the accurate position.}
  \label{fig:11}
\end{figure}

\begin{figure}[t]
  \centering
  \includegraphics[width=140mm]{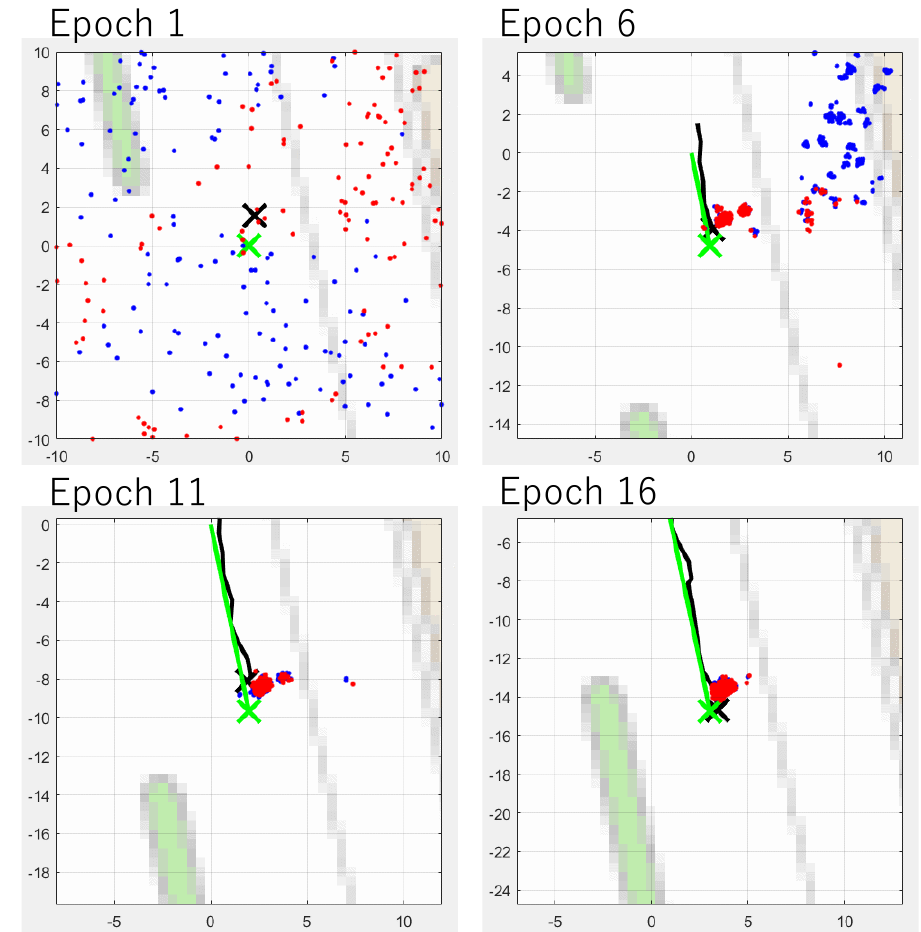} 
  \caption{Time series of the proposed particle filter processing. Particles are converged to the true position.}
  \label{fig:12}
\end{figure}

\begin{figure}[t]
  \centering
  \includegraphics[width=120mm]{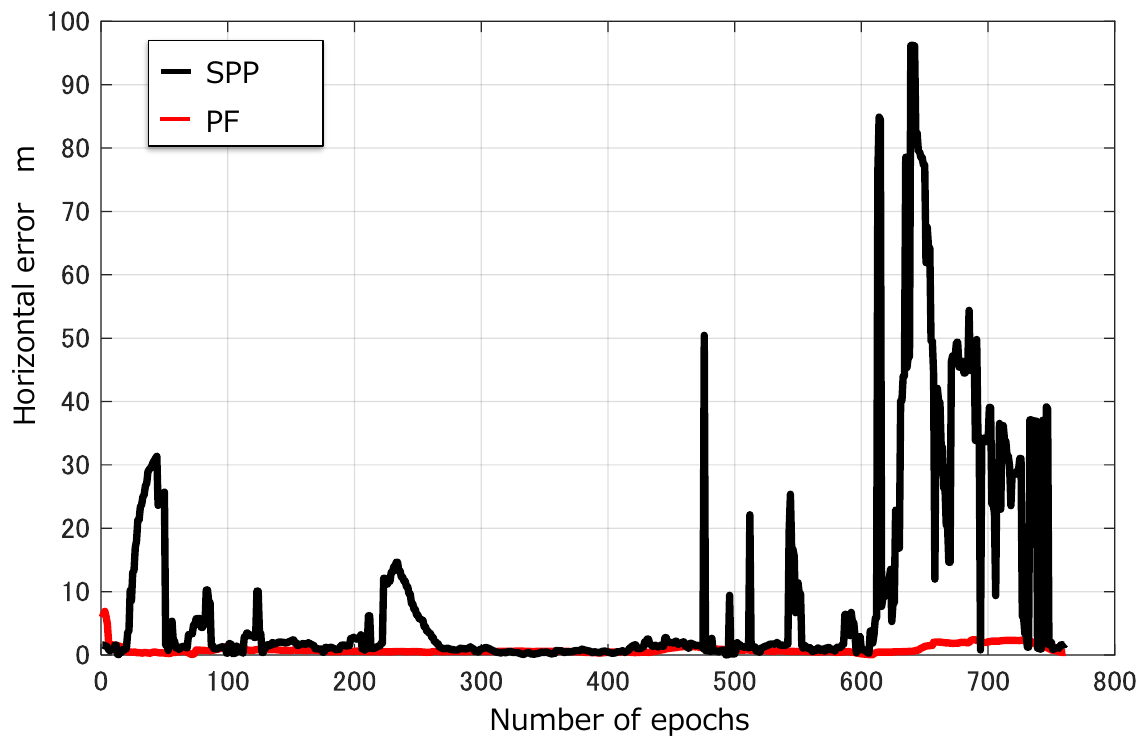} 
  \caption{Comparison of positioning error between SPP and the proposed method.}
  \label{fig:13}
\end{figure}

\begin{figure}[t]
  \centering
  \includegraphics[width=120mm]{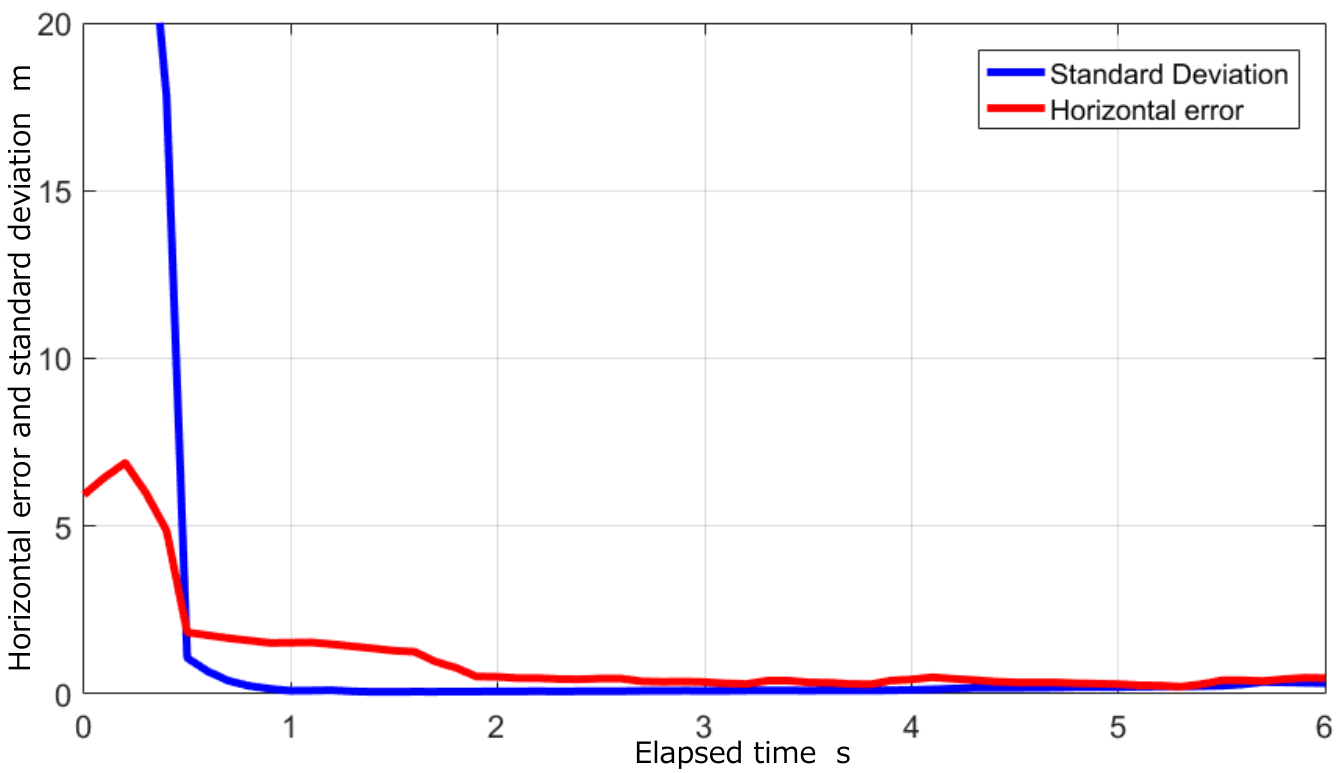} 
  \caption{Horizontal error and standard deviation using the proposed method.}
  \label{fig:14}
\end{figure}

We applied a particle filter to estimate the user position. Five hundred particles were used in this test. The particle filter calculations were performed during post-processing. The estimated user position is plotted in Figure 11. Compared with the conventional point positioning, the proposed technique can estimate the smooth trajectory.  Figure 12 shows a time series of the proposed particle filter processing. The blue and red points indicate the positions of the particles before and after the resampling step, respectively, over an interval of 0.5 s. In Figure 12, the distribution of particles, which had particles distributed in all regions prior to the resampling step, became more concentrated after resampling. After a few iterations of resampling, the particle distribution converged to the true position. The horizontal root mean square (RMS) of the localization error is plotted in Figure 13. Compared with the conventional point positioning, the horizontal RMS error of the proposed method decreased from 19.4 to 1.08 m. The time-series horizontal position error and standard deviation of the particle distribution are shown in Figure 14. After 2 s and iterated resampling, the distribution of the particles converged to within 1 m. The convergence time is fast because the proposed method integrates the pseudorange based GNSS positioning into the likelihood estimation process. These results show that our proposed method can be used to estimate the position with meter-scale accuracy in urban canyon environments, where conventional GNSS positioning results in large positioning errors.

\section{Conclusion}
We proposed a GNSS positioning method for urban environments based on a particle filter with NLOS GNSS signal detection without using any external data or sensors for the vehicles and mobile robots. The particle motion was estimated from the velocities computed by Doppler measurements and the proposed method is suitable for kinematic applications. Note that this technique also can be applied to the localization of a static object. We proposed a likelihood estimation method using the Mahalanobis distance between the particle location and new GNSS position solution using the subsets of the LOS measurement hypothesis determined from the pseudorange residuals. From our experimental results, we concluded that the proposed method can be used to correctly determine the particle likelihoods. Using the particle filter, the proposed method was used to perform accurate user positioning in urban canyons, where conventional GNSS positioning cause large positioning errors.

\bibliographystyle{tADR}
\bibliography{ar2018}

\end{document}